\documentclass[11pt,a4paper]{article}
\usepackage[hyperref]{eacl2021}
\usepackage{times}
\usepackage{latexsym}
\usepackage{graphicx}
\usepackage{multirow}
\usepackage{amsmath}
\usepackage{subfigure}

\usepackage{enumitem}
\usepackage{placeins}

\setlist[description]{style=unboxed}

\usepackage{microtype}

\aclfinalcopy % Uncomment this line for the final submission
\title{How to Evaluate a Summarizer: Study Design and Statistical Analysis for Manual Linguistic Quality Evaluation}

\author{Julius Steen \and Katja Markert\\
  Department of Computational Linguistics \\
  Heidelberg University\\
  69120 Heidelberg, Germany \\
  {\tt (steen|markert)@cl.uni-heidelberg.de}}
\date{}

\begin{document}
\maketitle

\begin{abstract}
Manual evaluation is essential to judge progress on automatic text summarization. However, we conduct a survey on recent summarization system papers that reveals little agreement on how to perform such evaluation studies. 
We conduct  two evaluation experiments on two aspects of summaries' linguistic quality (coherence and repetitiveness) to  compare Likert-type
and ranking annotations and show that best choice of evaluation method can vary from one aspect to another. 
In our survey, we also find that study parameters such as the overall number of annotators and distribution of annotators to annotation items are often not fully reported and that
subsequent statistical analysis ignores grouping factors arising from one annotator judging multiple summaries. Using our
evaluation experiments, we show
that the total number of annotators can have a strong impact on study power and that current statistical analysis methods
can inflate type I error rates up to eight-fold. In addition, we highlight that for the purpose of system comparison the current practice of eliciting
multiple judgements per summary leads to less powerful and reliable annotations given a fixed study budget.
\end{abstract}

\section{Introduction}

Current automatic metrics for summary evaluation 
have low correlation with human judgements on summary quality, especially for linguistic quality evaluation \citep{sumeval}. As a consequence, manual evaluation is still vital to properly compare the linguistic quality of summarization systems.

While the document understanding conferences (DUC) established a standard manual evaluation procedure  \citep{duc2005}, we conduct a  comprehensive survey of recent works in text summarization that reveals a wide array of different evaluation questions and methods in current use. Furthermore, DUC procedures were designed for a small set of expert judges, while current evaluation campaigns are often conducted by untrained crowd-workers.
The design of the manual annotation, specifically \textit{the overall number of annotators} as well as \textit{the distribution of annotators to annotation items}, has substantial impact on power, reliability and
type I errors of subsequent statistical analysis. However, most current papers (see Section~\ref{sec:survey}) do not consider the interaction of annotation design and statistical analysis.
We investigate the optimal annotation methods,  design and statistical analysis of summary evaluation studies, making the following contributions:

\begin{enumerate}
    \item We conduct a comprehensive survey on the current practices in manual summary evaluation in Section~\ref{sec:survey}. Often, important study parameters, such as 
    the total number of annotators, are not reported. In addition, statistical significance is either not assessed at all or with tests (t-test or one-way ANOVA) that  lead to inflated type I error in the presence of grouping factors \citep{barr-2014-mem}. In summarization evaluation, grouping factors arise  whenever 
    one annotator rates multiple summaries.
    
    \item We carry out annotation experiments for coherence and repetition.
    We use both Likert- and ranking-style questions on the output of four recent summarizers and reference summaries. %NewChange
    We show that ranking-style evaluations are more reliable and cost-efficient for coherence, similar to  prior findings by \citet{novikova-2018-rankme} and \citet{sakaguchi-2018-bounded_support}. However, on repetition, where many documents do not exhibit any problems, Likert outperforms ranking.

    \item Based on our annotation data, we perform Monte-Carlo simulations to %CHANGED 
    show the risk posed by ignoring grouping factors in statistical analysis and find up to eight-fold increases in type I errors when using standard significance tests. As an alternative, we propose to either use mixed effect models \cite{barr-2014-mem} for analysis or to design studies in such a manner that results can be aggregated into independent samples, amenable to simpler analysis tools.

    \item Finally, we show that the common practice of eliciting repeated judgements for the same summary leads to less reliable and powerful studies for  system-level comparison when compared to studies with the same budget but only one judgement per summary.
\end{enumerate}

Code and data for our experiments is available at \url{https://github.com/julmaxi/summary_lq_analysis}.

\section{Literature Survey}
\label{sec:survey}

\begin{table}
    \centering
    \begin{tabular}{|c|l|r|r|}
    \hline
&Category &    Pap. & Std. \\\hline
\multirow{7}{2cm}[-0.5em]{Evaluation Questions} & Overall             &     17 &          23 \\
& Content         &     45 &          65 \\
& Fluency         &     29 &          34 \\
& Coherence           &     10 &          11 \\
& Repetition          &     14 &          17 \\
& Faithfulness        &      6 &           8 \\
& Referential Clarity &      2 &           2 \\
& Other               &      8 &           9 \\
\hline
\multirow{7}{2cm}[-0.3em]{Evaluation Method}
&Likert   &     32 &          43 \\
&Pairwise &     10 &          14 \\
&Rank     &      9 &           9 \\
&BWS      &      6 &           9 \\
&QA       &      9 &          14 \\
&Binary   &      4 &           4 \\
&Other    &      2 &           2 \\\hline
\multirow{6}{2cm}[-0.6em]{Number of Documents in Evaluation} & $<20$   &      6 &          10 \\
&20-34 &     22 &          41 \\
&35-49 &      3 &           4 \\
&50-99 &     14 &          21 \\
&100   &     11 &          14 \\
&$>100$  &      4 &           4 \\\cline{2-4}
&\textit{not given} &      1 &           1 \\\hline

\multirow{6}{2cm}[-0.7em]{Number of Systems considered} & $<3$ &     13 &          20 \\
&3   &     17 &          23 \\
&4   &     16 &          23 \\
&5   &      6 &          10 \\
&$>5$  &     12 &          19 \\\cline{2-4}

&w/ Reference  &     16 &          25 \\
&w/o Reference &     45 &          70 \\\hline

\multirow{6}{2cm}[0.3em]{Number of Annotations per Summary} & 1     &      2 &           5 \\
&2-3   &     20 &          30 \\
&4-5   &     12 &          27 \\
&6-10  &      3 &           5 \\\cline{2-4}
&\textit{not given} &     23 &          28 \\\hline

\multirow{4}{2cm}{Overall Number of Annotators}
& 1-5   &     19 &          25 \\
& 6-10  &      3 &           3 \\
& $>10$   &      5 &           9 \\\cline{2-4}
&\textit{not given} &     32 &          58 \\\hline

\multirow{2}{2cm}{Annotator Recruitment} &Crowd &     25 &          49 \\
&Other &     35 &          46 \\\hline

\multirow{5}{2cm}{Statistical Evaluation}
& t-test      &      9 &          16 \\
& ANOVA       &      9 &          18 \\
& CI          &      4 &           6 \\
& Other/unspecified &      7 &           8 \\
& None          &     32 &          47 \\\hline
\end{tabular}
    \caption{Our survey for 58 system papers with 95 manual evaluation studies (2017-2019). We show numbers both for individual studies and per paper. As a paper may contain several studies with different parameters, counts in the paper column do not always add up.}
    \label{tab:survey_results}
\end{table}

We survey all summarization papers in ACL, EACL, NAACL, ConLL, EMNLP, TACL and the \textit{Computational Linguistics} journal in the years 2017-2019. 
We choose this timeframe as we are interested in current practices in summarization evaluation: 2017 marks the publication of the pointer generator network \citep{see-2017-pg}, which has been highly influential for neural summarization.
We focus our analysis on papers that present \textit{a novel system} for single- or multi-document summarization and take a single or multiple full texts as input and also output text (SDS/MDS). This allows us to concentrate on recommendations for human evaluation of newly developed summarization systems.\footnote{Excluded from the analysis are
sentence summarization or headline generation papers, although most of the points we make hold for 
their evaluation campaigns as well. Summarization evaluation papers that do not present a new system
but concentrate on sometimes large-scale system comparisons are discussed in the Related Work section instead. Lists of all included and excluded papers are given in Supplementary Material, which also
contains exact evaluation parameters per paper in a spreadsheet.}

Out of the resulting \textbf{105} SDS/MDS system papers, we identify all papers that conduct at least one new comparative system evaluation with human annotators for further analysis, leading to \textbf{58} papers in the survey.
The fact that this is only about half of all papers is troubling given that it has been recently demonstrated that current automatic evaluation measures such as ROUGE \citep{lin-2004-rouge} are not good at predicting summary scores for modern systems \citep{schluter-2017-limits, kryscinski-2019-neural, peyrard-2019-range}.

We  assess both \textit{what} studies ask annotators to judge, as well as \textit{how} they elicit and analyse  judgements. 
The survey was conducted by one of the authors: for most papers, the categories they fell into were obvious. For difficult cases (unclear specifications, papers that do not fit the normal mould)
the two authors discussed the categorisations. 
Survey results are given in Table~\ref{tab:survey_results}. Further details about the choices made in the survey, including category groupings/definitions and what is included under \textit{Other}, can be found in Appendix \ref{app:survey}.
As many papers conduct more than one human evaluation (for example on different corpora), we also list individual annotation studies (a total of 95).

Of the systems that do have human evaluation, many focus on \textit{content}, including
informativeness, coverage, focus, and relevance.
Where linguistic quality is evaluated, most focus on general questions about fluency/readability, with a smaller number of papers evaluating coherence and repetition.

In the rest of this section we focus on the three aspects of evaluation we cover in this paper:
How to elicit judgements, how these judgements are ana\-lysed statistically and how studies are designed.

\subsection{Methods}

The majority of evaluations is conducted using Likert-type judgements, with the second most frequent method being rank-based annotations, including pairwise comparison. Best-worst scaling (BWS)
is a specific type of ranking-oriented evaluation that requires annotators to specify only the first and last rank \citep{kiritchenko-2017-bws}. QA \citep{narayan-2018-ranking} is used for content evaluation only. This motivates us to compare both Likert and ranking annotations in Section~\ref{sec:reliability}.

\subsection{Statistical Analysis}

If a significance test is conducted, most papers analyse their data either using ANOVA or a sequence of paired t-tests. Both tests are based on the assumption that judgements (or pairs of judgements, in case of paired t-test) are sampled \textit{independently} from each other. However, in almost all studies, annotators give judgements on more than one summary from the same system.
Thus the resulting judgements are only independent if we assume that all annotators behave identically. Given that prior work \citep{gillick, amidei-2018-rethinking}, as well as our own reliability analysis in Section \ref{sec:reliability}, show that especially crowd-workers tend to disagree about judgements, this
assumption does not seem warranted.
As a consequence, traditional significance tests are at high risk of inflated type I error rates. This is well known in the broader field of linguistics \citep{barr-2014-mem}, but is disregarded in summarization evaluation. We show in Section~\ref{sec:significance} that this is a substantial problem for current summarization evaluations and suggest alternative analysis methods.

\subsection{Design}

Most papers only report the number of documents in the evaluation and the number of judgements \textit{per summary}. This, however, is not sufficient to describe the design of a study, lacking any indication about 
the overall number of annotators that made these judgements. A study with 100 summaries and 3 annotations per summary can mean 3 annotators did  all judgements in one extreme, or a study with 300 distinct annotators in the other. Only 26 of the 95 studies describe their annotation design in full, almost all of which use designs in which a small number of annotators judge all summaries. Only \textbf{6} of \textbf{49} crowdsourced studies report the full design.

We show in Section~\ref{sec:significance}  that a low total number of annotators aggravates type I error rates with improper statistical analysis. In Section~\ref{sec:design} we further show that with proper analysis, a low total number of annotators leads to less powerful experiments.
Almost all analysed papers choose designs with multiple judgements per summary. However, we show in Section~\ref{sec:power} that this --- for the purpose of system ranking --- leads to loss of reliability as well as power when compared to a study with the same budget and only one annotation per summary.

\section{Coherence and Repetition Annotation}

To elicit summary judgements for analysis, we conduct studies on two linguistic quality tasks. In the first, we ask annotators to judge the \textit{coherence} of the summaries, while in the second we ask for the \textit{repetitiveness} of the summary.
We select these two tasks over the more frequent \textit{Fluency} task as we found in preliminary investigations that many recent summarization systems already produce highly fluent text, making them hard to differentiate.
We do not evaluate \textit{Overall} and \textit{Content} as both require access to the input document, which differentiates these questions from the linguistic quality evaluation of the summaries.

For both tasks, we conduct one study using a seven-point  Likert-scale (\texttt{Likert}) and another using a ranking-based annotation method (\texttt{Rank}), where annotators  rank summaries for the same document from best to worst. Screenshots of the interfaces for both approaches and  full annotator instructions 
are given in Appendix \ref{app:instructions}.

\paragraph{Corpus and Systems.} Mirroring a common setup (see Section~\ref{sec:survey}), 
we select four abstractive summarization systems and the reference summaries (\texttt{ref}) for analysis.

\begin{itemize}
    \item The pointer generator summarizer (\texttt{PG}) \citep{see-2017-pg}, which is still often used as a baseline for abstractive summarization 
    \item The abstractive sentence rewriter (\texttt{ASR}) of \citet{gehrmann-2018-bottomup}, which is a strong summarization system that does not rely on external pretraining for its generation step %CHANGE
    \item \texttt{Seneca} \citep{sharma-2019-entity}, a system that combines explicit modelling of coreference information with an external coherence model 
    \item \texttt{BART} \citep{lewis-2019-bart}, a transformer network that achieves SotA on CNN/DM.
\end{itemize}

We randomly sample 100 documents from the popular CNN/DM corpus \citep{hermann} with corresponding summaries from all systems to form the item set for all our studies.

\paragraph{Study design.}
We ensure a sufficient total number of annotators
by using a block design.
We separated our corpus into 20 blocks of 5 documents and included all 5 summaries for each document in the same block, which results in 5 $\times$ 5 = 25 summaries per block.

\begin{figure}
    \centering
    \includegraphics[width=0.3\linewidth]{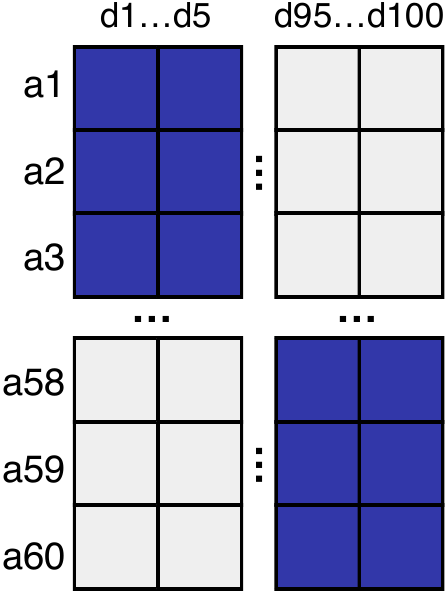}
    \caption{Schematic representation of our study design. Rows represent annotators, columns documents. Each blue square corresponds to a judgement of the summaries of all five systems for a document. Every rectangular group of blue squares forms one block.}
    \label{fig:design}
\end{figure}

All items in a block were judged by the same set of three annotators. No annotator was allowed to judge more than one block. This results in a total of 3 $\times$ 20 = 60 annotators and 1500 judgements per task. Figure~\ref{fig:design} shows a schematic overview of our design, which balances the need for a large enough annotator pool with a sufficient task size to be worthwhile to annotators.

We recruited native English speakers from the crowdsourcing platform Prolific\footnote{\url{prolific.com}} and carefully adjusted the reward to be no lower than \pounds $7.50$ per hour based on pilot studies. 
Summaries (or sets of summaries for \texttt{Rank}) within a block were presented in random order.

\section{Ranking vs. Likert}

\begin{table*}
    \centering
    \begin{tabular}{|l|c|c||c|c|c|}
    \hline
 System       & \texttt{Likert (Coh)} & \texttt{Rank (Coh)} & \texttt{Likert (Rep)} & \texttt{Rank (Rep)} \\\hline
    
\texttt{BART} & $5.25^{(1)}$ & $1.73^{(1)}$ & $5.85^{(2/3)}$ & $2.88^{(2/3/4)}$ \\
\texttt{ref} & $4.33^{(3/4)}$ & $3.31^{(3/4)}$ & $6.14^{(1/2)}$ & $2.41^{(1/2)}$\\
\texttt{ASR} & $4.17^{(3/4)}$ & $3.17^{(3/4)}$ & $4.88^{(4/5)}$ & $3.51^{(4/5)}$ \\
\texttt{PG} & $4.81^{(2)}$ & $2.68^{(2)}$ & $5.63^{(3)}$ & $2.92^{(3/4)}$ \\
\texttt{seneca} & $3.52^{(5)}$ & $4.11^{(5)}$ & $5.16^{(4/5)}$ & $3.27^{(3/4/5)}$
\\
\hline
    \end{tabular}
    \caption{Results of our annotation experiment. Numbers in brackets indicate rank for a system for a given annotation method. Multiple ranks in the brackets indicate systems at these ranks are not statistically significantly different $(p \geq 0.05$, mixed-effects ordinal regression).} 
    \label{tab:results}
\end{table*}

Table~\ref{tab:results} shows the average Likert scores and the average rank for all systems, tasks and annotation methods. We use mixed-effect ordinal regression to identify significant score differences (see Section~\ref{sec:significance} for details).
Both annotation methods provide  compatible system rankings for the two tasks, though for the repetition task both methods struggle to differentiate between systems.
If we were interested in the true ranking, we could conduct a power analysis given some effect size of interest and  elicit additional judgements to improve the ranking.
However, as we are concerned with the \textit{process} of system evaluation and not the evaluation itself, we do not conduct any further analysis.

In the remainder of this section, we focus on the reliability of the two methods as well as
their cost-effectiveness.

\subsection{Reliability}
\label{sec:reliability}

\begin{table}
    \centering
    \begin{tabular}{|l|c|c|c|}
    \hline
       System & $\alpha$  & SHR \\\hline
       Coh: \texttt{Likert} & 0.22 & 0.96  \\
       Coh: \texttt{Rank} & 0.43 & 0.98 \\
       \hline
       Rep: \texttt{Likert} & 0.27 & 0.95 \\
       Rep: \texttt{Rank} & 0.18 & 0.91 \\

    \hline
    \end{tabular}
    \caption{Krippendorffs $\alpha$ with ordinal level of measurement and Split-Half-Reliability for both annotation methods on the two tasks.}
    \label{tab:reliability}
\end{table}

Traditionally, reliability is computed by chance-adjusted agreement on individual instances. However, for NLG evaluation, \citet{amidei-2018-rethinking} argue that a low agreement often reflects variability in language perception. Additionally, we are not interested in individual \textit{document scores}, but in whether independent runs of the same study would result in consistent \textit{system scores}. In Table~\ref{tab:reliability} we thus report split-half reliability (SHR) in addition to  Krippendorffs $\alpha$ \citep{krippendorff-alpha}.
To compute SHR, we randomly divide judgements into two groups that share neither annotators nor documents, i.e. two independent runs of the study. We then compute the correlation\footnote{We use the Pearson correlation implementation of scipy \citep{scipy}.} between the system scores in both halves. The final score is the average correlation after 1000 trials.

Though agreement on individual summaries is relatively low for all annotation methods, studies still arrive at consistent system scores when we average over many annotators as demonstrated by the SHR. This reflects similar observations made by \citet{gillick}.

\begin{figure}
    \centering
    \includegraphics[width=\linewidth]{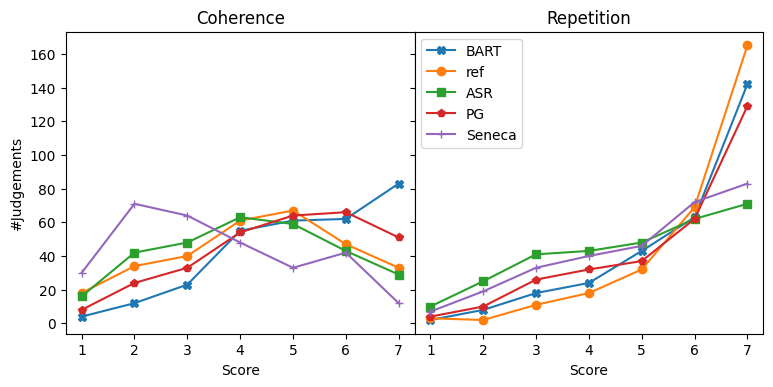}
    \caption{Score distribution of \texttt{Likert} for both tasks. Each data point shows the number of times a particular score was assigned to each system.}
    \label{fig:scoredist}
\end{figure}

We find that on coherence, \texttt{Rank} is more reliable than \texttt{Likert}, though not on repetition.
An investigation of the \texttt{Likert} score distributions for both tasks in Figure~\ref{fig:scoredist}
shows that coherence scores are relatively well differentiated whereas a majority of repetition judgements give the highest score of $7$, indicating no repetition at all in most summaries.
We speculate overall agreement suffers, because ranking summaries with similarly low level of repetition (and not allowing ties) is potentially arbitrary.\footnote{This is supported by feedback we received from annotators that the summaries were difficult to rank as they mostly avoided repetition well.}

%must add up to 300 right?

\subsection{Cost-efficiency}

\begin{figure}
    \centering
    \includegraphics[width=\linewidth]{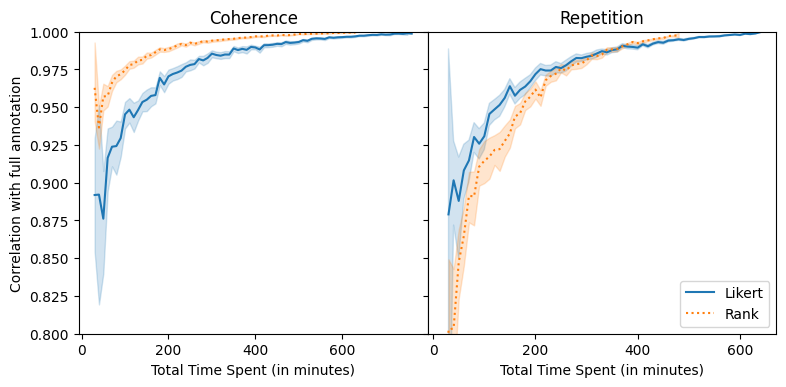}
    \caption{Time spent on annotation (in minutes) vs. correlation with the full-sized score. We gather annotation times in buckets with a width of ten minutes and show the 95\% confidence interval for each bucket.}
    \label{fig:timespower}
\end{figure}

While more reliable annotation methods allow for fewer annotations, the cost of a study is ultimately determined by the work-time that needs to be invested to achieve a reliable result.
To investigate this, we randomly sample between 2 and 19 blocks from our annotations and compute the total time annotators spent to complete each sample. We  also compute the Pearson correlation of the system scores in each sample with the scores on the full annotation set. We  relate time spent to similarity between sample and full score  in Figure~\ref{fig:timespower}.

For coherence, \texttt{Rank} is more efficient than \texttt{Likert}. On repetition, the lower reliability of \texttt{Rank} also results in lower efficiency. However, with additional annotation effort, reliability becomes on-par with \texttt{Likert}. This is a consequence of the overall lower annotator workload for \texttt{Rank}.

\section{Statistical Analysis and Type I Errors}
\label{sec:significance}

The two most common significance tests in summarization studies, ANOVA and t-test (see Table~\ref{tab:survey_results}), both assume judgements (or pairs of judgements, in the case of t-test) are independent.
This is, however, not true for most study setups as a single annotator typically judges multiple summaries and multiple summaries are generated from the same input document. Both documents and annotators are thus grouping factors in a study that must be taken into account by the statistical analysis.
Generalized mixed effect models \citep{barr-2014-mem} offer a solution but have, to the best of our knowledge, not been used in summarization evaluation at all.
We choose a mixed effect ordered logit model to analyse our Likert data for both tasks.\footnote{We do not include Rank data as the ordinal regression model does not generate ranks.} We will show that traditional analysis
methods have a substantially elevated risk of type I errors, i.e. differences between systems found in manual analysis might be overstated.

\paragraph{Method.}
The ordered logit model we employ can be described as follows:

\begin{align*}
    \mathit{logit}(P(Y \leq c)) \\
    = \mu_c - (X\beta + Z_{a}u_{a} + Z_{d}u_{d}) \\
\end{align*}

where $P(Y \leq c)$ is the probability that the score of a summary is at most $c$. $\mu_c$ is the threshold coefficient for level $c$, $\beta$ is the vector of fixed effects and $u_a, u_d$ are the vectors of annotator- and document-level random effects respectively, where $u_a, u_d$ are both drawn from normal distributions with mean $0$. Finally, $X, Z_a, Z_d$ are design matrices for fixed and random effects. 
As the only fixed effect, we use a dummy-coded variable indicating the system that has produced the summary, with \texttt{ref} as the reference level.
We estimate both random intercepts and slopes for both documents and annotators following advice of \citet{barr-2014-mem} to always specify the maximal random effect structure.
In practical terms this means that we allow annotators to both differ in how harsh or generous they are in their assessment, as well as in which system they prefer.
Similarly, we allow system performance to vary per-document, leading to both generally higher or lower scores, as well as different system rankings per document.

We fit all models using the \texttt{ordinal} R-package \citep{ordinal-r} and compute pairwise contrasts between the parameters estimated for each system using the \texttt{emmeans}-package \citep{emmeans} with Tukey-adjustment. %CHANGED

To demonstrate the problem of ignoring the grouping factors, we can now sample artificial data from the model distribution and try to analyse it with inappropriate tests. This Monte-Carlo simulation is similar to the more general analysis of \citet{barr-2014-mem}.

We set $\beta$ to 0 so all systems perform equally well on the population level and only keep the (zero-mean) document and annotator effects in the model.
The false-positive rate of statistical tests on this artificial data should thus be no higher than the significance level.
We then repeatedly apply both the t-test and the approximate randomization test (ART) \citep{art}, a non-parametric test, to samples drawn from the model and determine the type I error rate at $p < 0.05$. We set the number of documents to $100$ and demand $3$ judgements per summary to mirror a common setup in manual evaluation. 
We  then vary the total number of annotators between $3$ and $300$ by changing how many summaries a single annotator judges.

\paragraph{Results.}

\begin{figure}
    \centering
    \includegraphics[width=\linewidth]{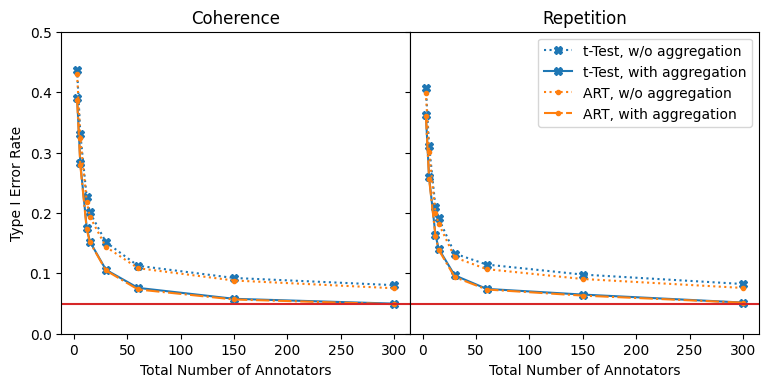}
    \caption{Relation of type I error rates at $p < 0.05$ to the total number of annotators for different designs, all with 100 documents and 3 judgements per summary. We conduct the experiment with both the t-test and approximate randomization test (ART). We show results both with averaging results per document and without any aggregation. We run 2000 trials per design. The red line marks the nominal error rate of $0.05$.}
    \label{fig:error_rates}
\end{figure}

We report results given the model estimated for \texttt{Likert} in Figure~\ref{fig:error_rates}. 
Ignoring the dependencies between samples leads to inflated type I error rates, whether using the t-test or the ART.
This is especially severe
when only few annotators judge the whole corpus. 
In the extreme case with only three annotators in total, the null-hypothesis is rejected in about 40\% of trials at a significance level of $0.05$ in both tasks. Even our original design with 60 annotators still sees an increase of the type I error rate by about $3\%$.
Only if every annotator judges a single document and annotations are averaged per document, samples are independent and thus the real error is at the nominal $0.05$ level. This design, however, is unrealistic given that annotators must be recruited and instructed.

We suggest two solutions to this problem: Either use mixed effect models or aggregate the judgements so samples become independent. This allows the assumptions of simpler tools such as ART to be met. In our study, we could average judgements in every block to receive independent samples. This is only possible, however, if the design of the study considers this problem in advance:
a crowd-sourcing study that allows annotators to judge as many samples as they like is unlikely to result in such a design.

\section{Study Design and Study Power}
\label{sec:design}

When conducting studies for system comparison, we are interested in maximizing their power to detect differences between systems. For traditional analysis, the power is ultimately determined by the number of documents (or judgements, when no aggregation takes place) in the study. However, when analysis takes into account individual annotators, power becomes additionally dependent on the total number of annotators  and how evenly they participated in the study. This gives additional importance to the design of evaluation studies. In this section, we thus focus on how to optimize studies for power and reliability.

We first show that for well-powered experiments, we need to ensure that a sufficient total number of annotators participates in a study. In the second part of this section, we will then demonstrate studies can improve their power by not eliciting multiple judgements per summary.

\subsection{Overall Number of Annotators}

To demonstrate the difference in power caused by varying the total number of annotators in a study, we determine the power for a design with the same total number of documents and judgements per document but different total numbers of annotators.

We run the experiment both with regression and ART with proper aggregation of dependent samples as described in Section~\ref{sec:significance}. We refer to the latter as \texttt{ARTagg} to differentiate it from normal ART.

For each design we repeatedly sample artificial data from the \texttt{Likert} model and apply both tests to the data.
The process is the same as in Section~\ref{sec:significance}  except we do not set $\beta$ to zero and count acceptances of the null-hypothesis.\footnote{As this is an observed power analysis it probably overestimates the power of our analysis for the true effect. 
The analysis is thus only useful to compare designs under our best estimate of actual effect sizes.}

We again set the number of documents to 100 and the number of repeat judgements to 3 and vary the total number of annotators between 3 and 75 by varying the number of blocks between 1 and 25. We test for power at a significance level of $0.05$.

\begin{figure}
    \centering
    \includegraphics[width=\linewidth]{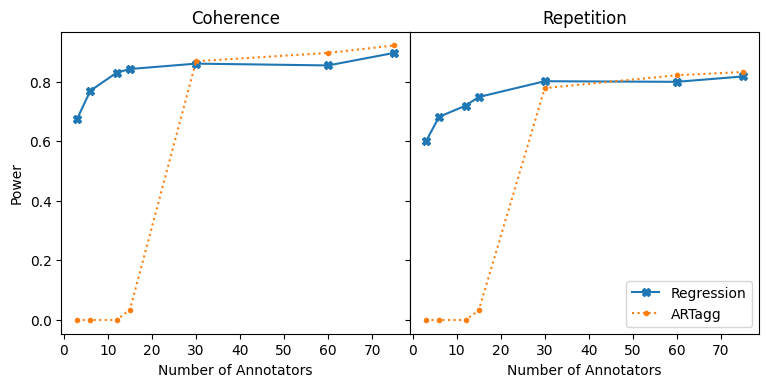}
    \caption{Power for 100 documents and 3 judgements per summary with different number of total annotators.}
    \label{fig:annopower}
\end{figure}

Figure~\ref{fig:annopower} shows how power drops sharply when only few annotators take part in the study. This is in line with the theoretical analysis of \citet{Judd} that shows that the number of participants is crucial for power when analysing studies with mixed effect models. The drop is worse for \texttt{ARTagg} as fewer annotators mean fewer independent blocks and thus a lack of datapoints for the analysis.

\subsection{Annotator Distribution}
\label{sec:power}

Most studies elicit multiple judgements per summary, following best practices in NLP for corpus design \citep{carletta}.  While this leads to better judgements per \textit{document}, the goal of 
many summarization evaluations is a per \textit{system} judgement.

For this kind of study, \citet{Judd} show that for mixed models that include both annotator and target (in our case, input document) effects, a design where targets are \textit{nested} within annotators, i.e. every annotator has its own set of documents, is always more powerful than one where they are (partially) \textit{crossed} with annotators, i.e. a study with multiple annotations per summary, \textit{given the same total number of judgements}. In fact, power could be maximized by having each annotator judge the summaries for only a single, unique document. However, this is usually not realistic due to the fixed costs of annotator recruitment and instruction.  % NewChange
We demonstrate  on our dataset how both reliability and power are affected by nested vs. crossed design.

\begin{figure}
    \centering
    \includegraphics[width=\linewidth]{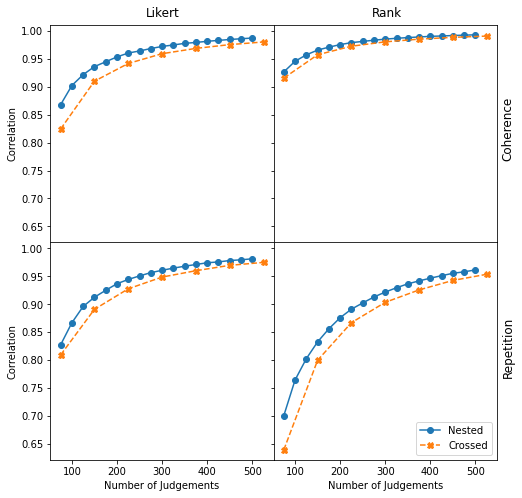}
    \caption{Reliabilities of nested vs. crossed designs for \texttt{Rank} and \texttt{Likert} for both tasks.}
    \label{fig:multiannotator}
\end{figure}

To compare reliability, we randomly sample both nested and crossed designs from our full study and then compute the Pearson correlation of the system scores given by this smaller annotation set with the system scores given by the full study.
As shown in Figure~\ref{fig:multiannotator},
nested samples are always at least as good and mostly better at approximating the results of the full annotation compared to a crossed sample with the same annotation effort.

We also conduct a power analysis for regression and \texttt{ARTagg} comparing nested and crossed designs.
We again turn to Monte-Carlo simulation on the \texttt{Likert} models and sample nested and crossed designs with the same total number of judgements (i.e. the same cost). We keep the block size constant at $5$ and vary the number of annotators between $3$ and $60$. For nested designs, we drop the document-level random effects from the ordinal regression, as document is no longer a grouping factor in nested designs.

\begin{figure}
    \centering
    \includegraphics[width=\linewidth]{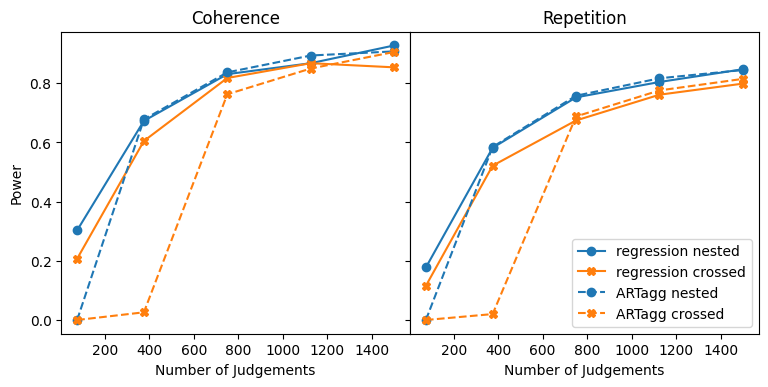}
    \caption{Power for $p < 0.05$ of nested and crossed designs for \texttt{ARTagg} and regression. X-axis shows the number of judgements elicited, Y-axis the power-level.}
    \label{fig:poweranalysis}
\end{figure}

Figure~\ref{fig:poweranalysis} shows that nested designs always have a power advantage over crossed designs, especially when few judgements are elicited. We also find that ART can be used to analyse data without loss of power when there are enough independent blocks. This might be attractive as ART is less computationally expensive than  ordinal regression.

\section{Related Work}

Human evaluation has a long history in summarization research. This includes work on the correlation of automatic metrics with human judgements \cite{lin-2004-rouge, liu-2008-correlation, graham-2015-reevaluating, peyrard-2017-principled, gao-2019-automated, sun-2019-feasibility, xenouleas-2019-sumqe, zhao-2019-moverscore, sumeval, gao-2020-supert} and improving the efficiency of the annotation process \citep{pyramid,hardy-2019-highres, shapira-etal-2019-crowdsourcing}. 
The impact of annotator inconsistency on system ranking has been studied both by \citet{owczarzak-2012-assessment} and \citet{gillick}. To the best of our knowledge, we are the first to
investigate the implications of annotator variance on the statistical analysis and the design in summarization system comparison studies.

For general NLG evaluation, \citet{van-der-lee-2019-best-practices} establish best practices for evaluation studies. We extend on their advice by conducting experimental studies specifically for summary evaluation. In addition, we show the importance of study design and consideration of annotator-effects in analysis on real world data. The advice of \citet{mathur-2017-sequence} regarding annotation sequence effects should be taken into account in addition to our suggestions.

\paragraph{Method Comparison.} Ranking has been shown to be effective in multiple NLP-tasks \citep{kiritchenko-2017-bws, zopf-2018-estimating}, including NLG quality evaluation \citep{novikova-2018-rankme}. In this work we confirm this for coherence evaluation, although we find  evidence that ranking is less efficient on repetition, where many documents do not exhibit any problems. We also add the dimension of annotator workload as a primary determinant of cost to the analysis of the comparison.

 Multiple methods have been suggested to reduce study cost by sample selection \citep{sakaguchi-2014-mt, novikova-2018-rankme, sakaguchi-2018-bounded_support, liang-2020-beyond} or integration with automatic metrics \citep{chaganty-2018-debias}. These efforts complement ours, as care still needs to be taken in analysis and study design.

Recently, rank-based magnitude estimation has been shown to be a promising method for eliciting judgements in NLG tasks and offers a combination of ranking and rating approaches \citep{novikova-2018-rankme, santhanam-shaikh-2019-towards}.
However, it has not yet found widespread use in the summarization community. While magnitude estimation has been shown to reduce annotator variance, our advice regarding experimental design and grouping factors in statistical analysis applies to this method as well, as annotators can still systematically differ in which systems they prefer.

\paragraph{Statistical analysis.} With regard to statistical analysis of experimental results, \citet{dror-2018-hitchhikers} give advice for hypothesis testing in NLP. However, they do not touch on the problem of dependent samples. \citet{rankel-etal-2011-ranking} analyse TAC data and show the importance of accounting for input documents in statistical analysis of summarizer performance and suggest the use of the Wilcoxon signed rank test for analysis.
\citet{azer-2020-bayes} argue that p-values are often not well understood and advocate bayesian methods as an alternative. While the analysis in our paper is frequentist, the mixed effect model approach can also be integrated into a bayesian framework. \citet{kulikov-etal-2019-importance} model annotator bias in such a framework but do not account for differences in annotator preferences.
In work conducted in parallel to ours, \citet{card-etal-2020-little} show that many human experiments in NLP underreport their experimental parameters and  are underpowered, including Likert-type judgements. Their simulation approach to power analysis is very similar to our experiments.  In addition to their analysis, we show that ignoring grouping factors in statistical analysis of human annotations leads to inflated type I error rates. We also show that power can be increased by choosing nested over crossed designs with the same budget.
The problem of underpowered studies has also been tackled outside of NLP by \citet{brysbaert-2019-annotators}.

For psycholinguistics, \citet{barr-2014-mem} demonstrate how generalizability of results is negatively impacted by ignoring grouping factors in the analysis. Mixed effect models have found use in NLP before \citep{green-2014-human, cagan-2017-data, karimova-2018-study, kreutzer-2020-correct}, but to the best of our knowledge they have not been used in summary evaluation.

\section{Conclusion}

We  surveyed the current state of the art in manual summary quality evaluation and investigated methods, statistical analysis and design of these studies. We distill our findings into the following guidelines for manual summary quality evaluation:

\paragraph{Method.} Both ranking and Likert-type annotations are valid choices for quality judgements. However, we present preliminary evidence that the optimal choice of method is dependent on task characteristics: If many summaries are similar for a given aspect, Likert may be the better option.

\paragraph{Analysis.} Analysis of elicited data should take into account variance in annotator preferences to avoid inflated type I error rates. We suggest the use of mixed effect models for analysis that can explicitly take into account grouping factors in studies. Alternatively, traditional tests can be used with proper study design and aggregation.

\paragraph{Study Design.} Study designers should control the number of annotators and how many summaries each individual annotator judges to ensure sufficient study power. Additionally, to ensure reliability of results, studies should report the design and the total number of annotators in addition to the number of documents and repeat judgements. Studies with repeat judgements on the same summary do not provide any advantage for system comparison and are less reliable and powerful than nested studies of the same size.

We hope that these findings will help researchers plan their own evaluation studies by allowing them to allocate their budget better. We also hope that our findings will encourage researchers to take more care
in the statistical analysis of results. This prevents misleading conclusions due to ignoring the effect of differences in annotator behaviour.

\section*{Acknowledgements}

We would like to thank Stefan Riezler for many fruitful discussions about the applications of mixed effect models.

  \bibliographystyle{acl_natbib}
  \bibliography{bib}

\appendix

\section{Interface Screenshots} \label{app:instructions}

\begin{figure*}
\subfigure[\texttt{Likert} - Coherence]{\label{fig:intro_likert_s}\includegraphics[width=75mm]{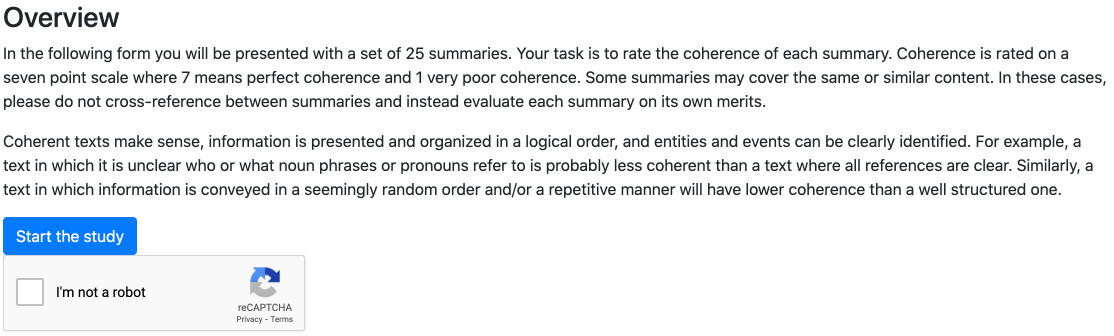}}
\subfigure[\texttt{Rank} - Coherence]{\label{fig:intro_rank}\includegraphics[width=75mm]{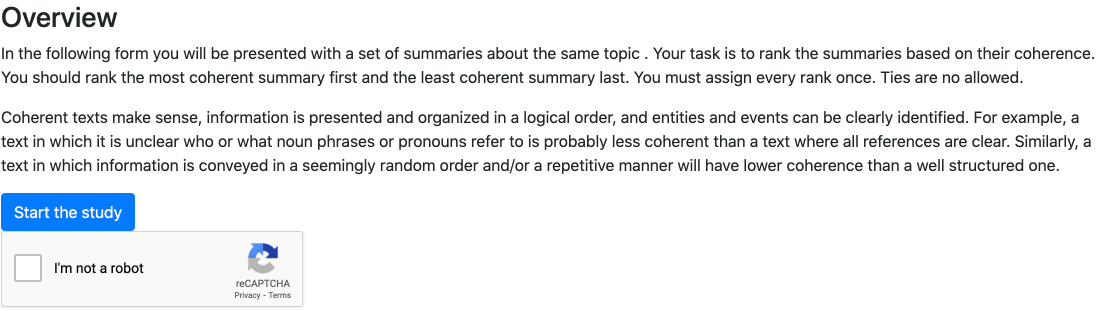}}

\subfigure[\texttt{Likert} - Repetition]{\label{fig:intro_likert_s_rep}\includegraphics[width=75mm]{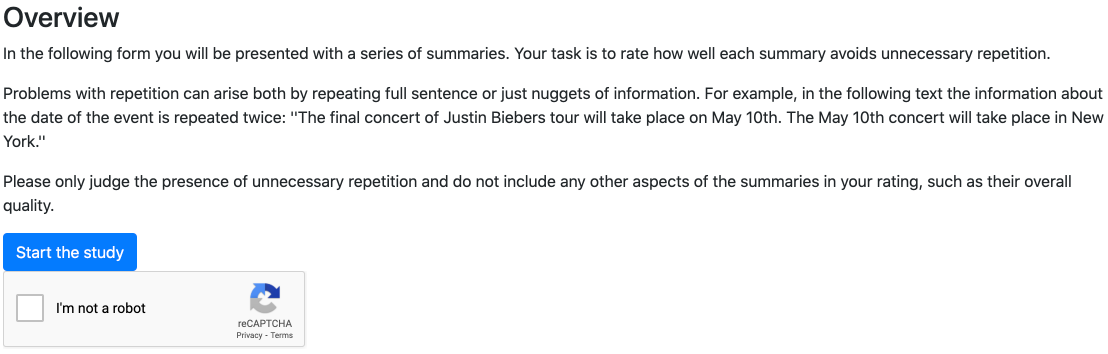}}
\subfigure[\texttt{Rank} - Repetition]{\label{fig:intro_rank_rep}\includegraphics[width=75mm]{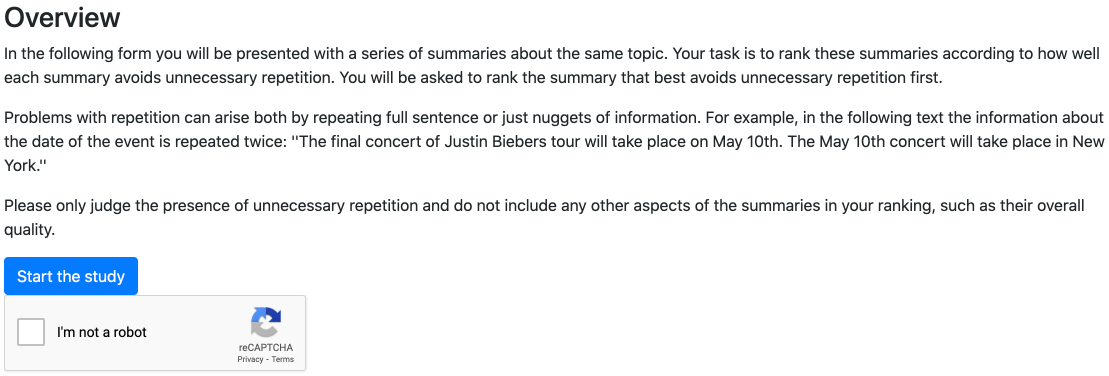}}

\caption{Screenshots of the Annotator Instructions.}
\label{fig:screenshots_intro}
\end{figure*}

\begin{figure*}
\subfigure[\texttt{Likert} - Coherence]{\label{fig:form_likert_s}\includegraphics[width=75mm]{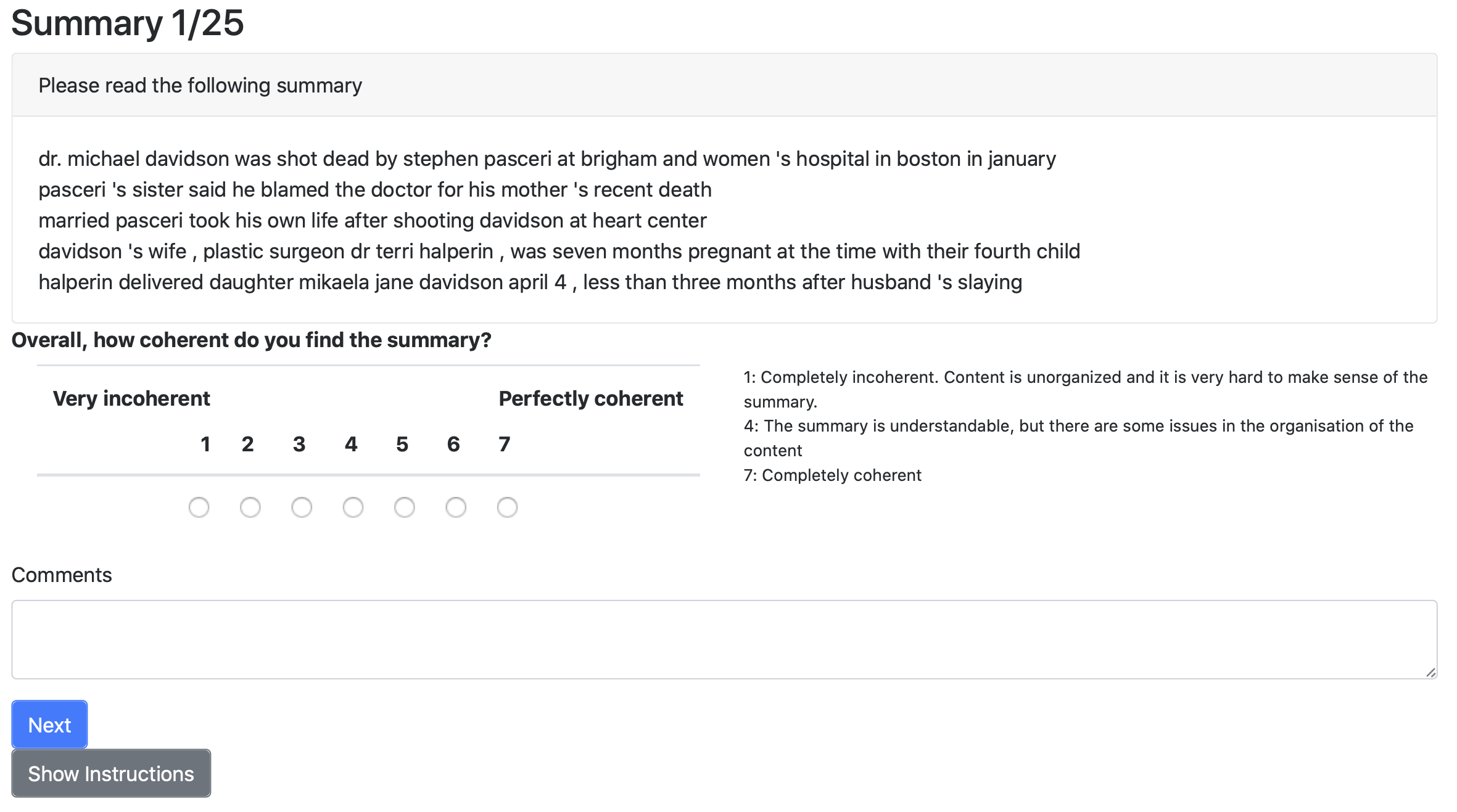}}
\subfigure[\texttt{Rank} - Coherence]{\label{fig:form_rank}\includegraphics[width=75mm]{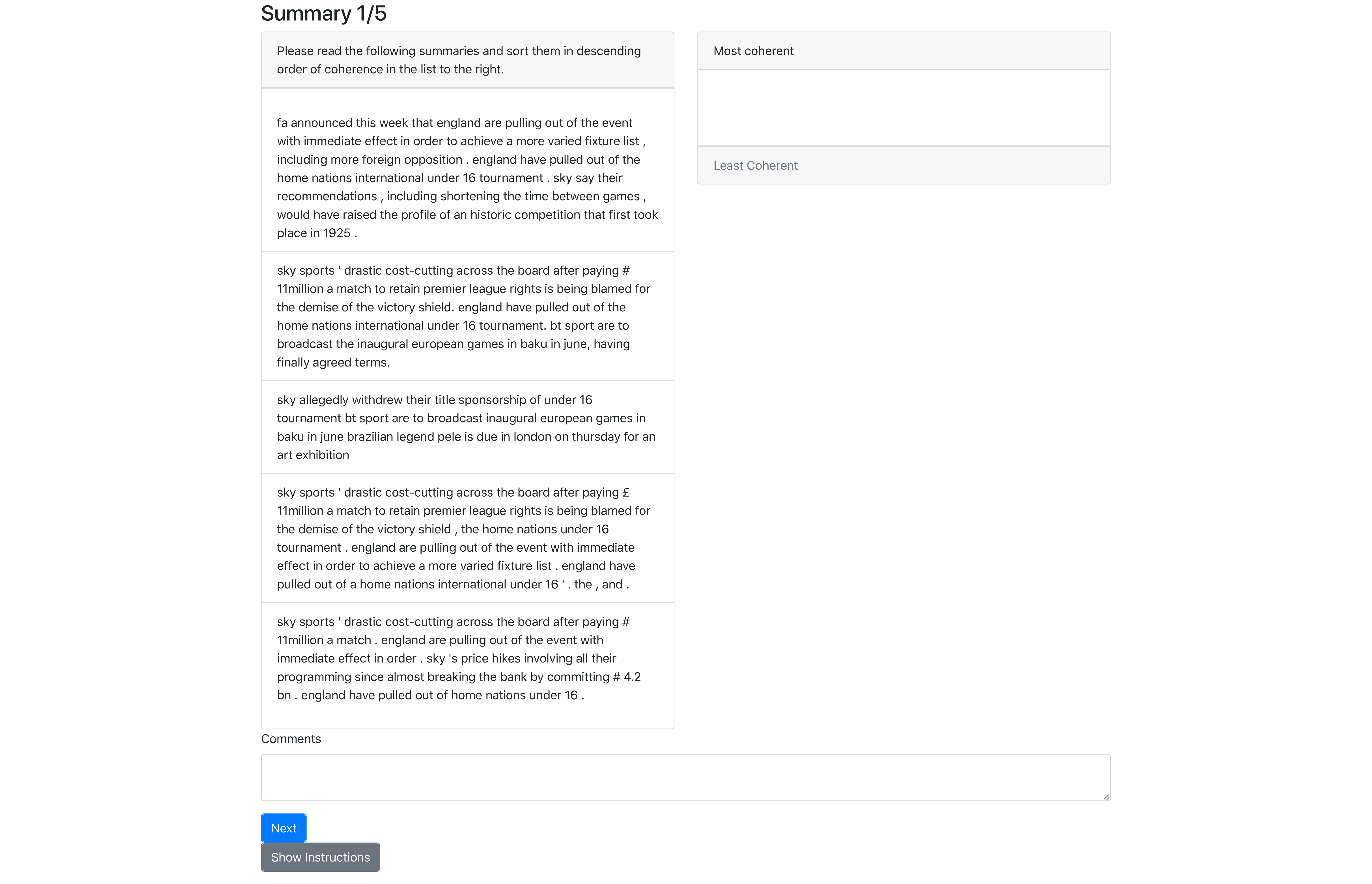}}

\subfigure[\texttt{Likert} - Repetition]{\label{fig:form_likert_s_rep}\includegraphics[width=75mm]{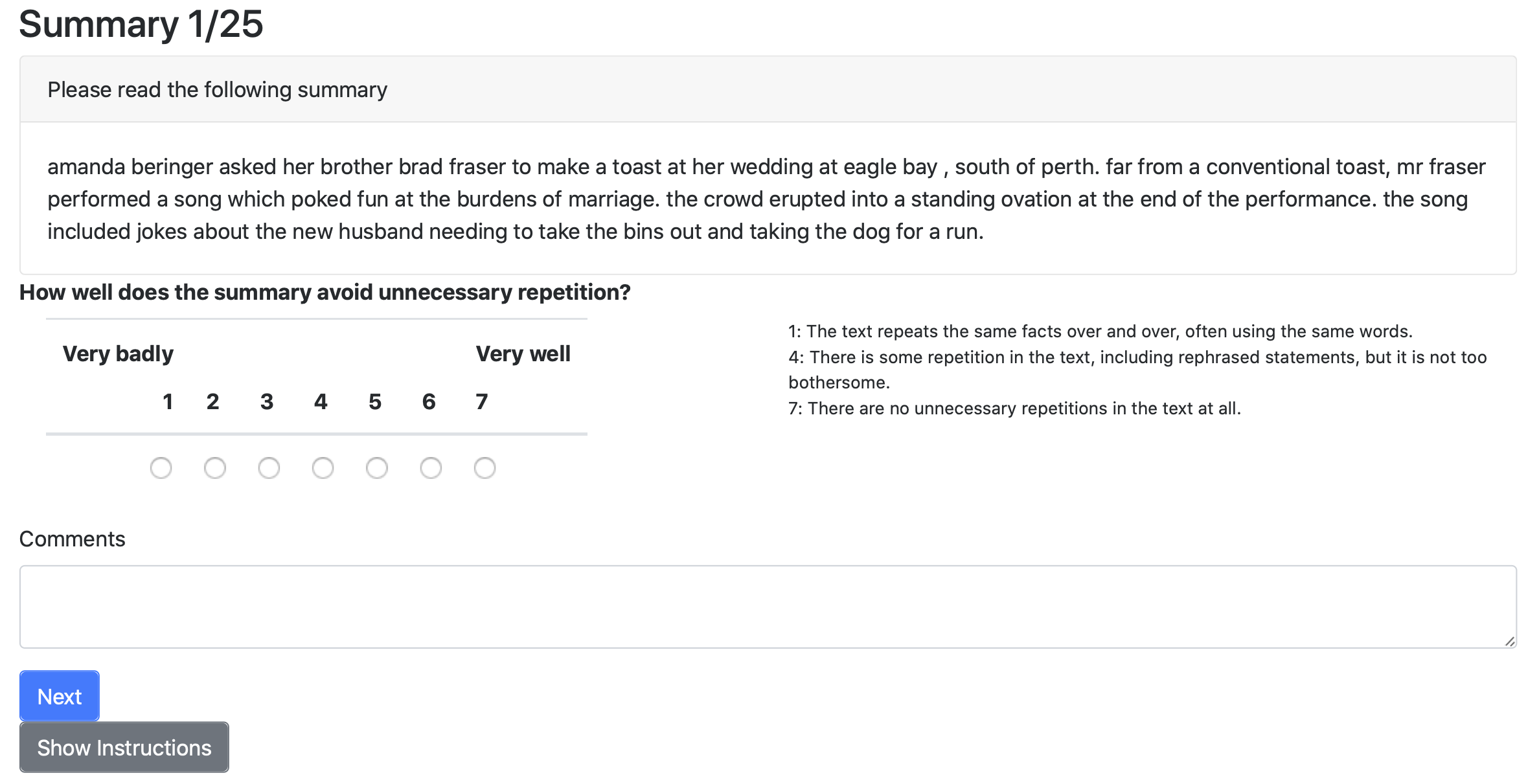}}
\subfigure[\texttt{Rank} - Repetition]{\label{fig:form_rank_rep}\includegraphics[width=75mm]{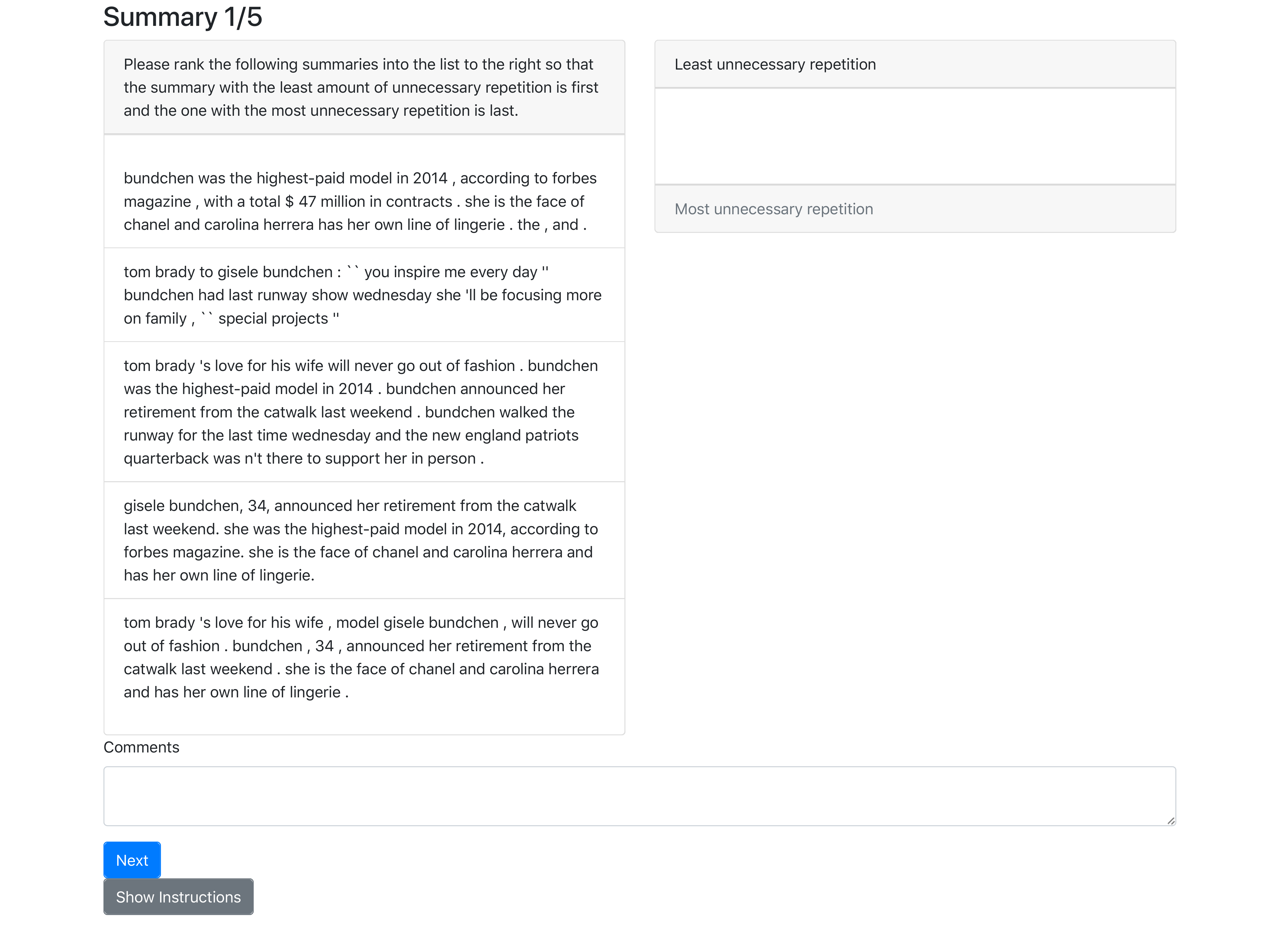}}

\caption{Screenshots of the Annotation Interfaces.}
\label{fig:screenshots}
\end{figure*}

We show screenshots of the instructions for both annotation methods and tasks in Figure~\ref{fig:screenshots_intro} and interfaces in Figure~\ref{fig:screenshots}.

\section{Survey} \label{app:survey}

\subsection{Categories}

While most categories are self-explanatory, we elaborate on some of the decisions we made during the survey in this section.

\paragraph{Evaluation Questions.} We allow a single study to include multiple evaluation questions,  as long as all questions are answered by the same annotators and use the same method. We make no distinction between informativeness, coverage, focus and relevance and summarize them under \textit{Content}. Similarly, we summarize fluency, grammaticality and readability under \textit{Fluency}. \textit{Other} includes:
\begin{itemize}
    \item One study with a specialized set of evaluation questions evaluating the usefulness of a generated related work summary % wang-etal-2018-neural-related
    \item One study of \textit{polarity} in a sentiment summarization context % angelidis-lapata-2018-summarization
    \item One study where annotators were asked to identify the aspect a summary covers in the context of review summarization % frermann-klementiev-2019-inducing
    \item Two studies evaluating formality and \textit{meaning similarity} of reference and system summary % chawla-etal-2019-generating
    \item One study evaluating diversity %
    \item One study conducting a Turing test % shen-etal-2019-improving
    \item One study asking paper authors whether they would consider a sentence part of a summary of their own paper. % erera-etal-2019-summarization
    \item One study evaluating structure and topic diversity
\end{itemize}

\paragraph{Evaluation Method.} \textit{Binary} includes any task with a yes/no style decision, while \textit{pairwise} includes any method in which two systems are ranked against each other. \textit{Other} includes
\begin{itemize}
    \item The aspect identification task mentioned above
    \item One study in which participants selected a single best summary out of a set of summaries
\end{itemize}

\paragraph{Annotator Recruitment.} \textit{Other} includes any recruitment strategy that does not rely on crowdsourcing. This includes cases in which the recruitment was not specified, students, experts, the authors themselves and various kinds of volunteers.

\paragraph{Statistical Evaluation.} \textit{Other/unspecified} includes
\begin{itemize}
    \item Four studies which reported statistical significance without reporting the test used
    \item Two studies using the approximate randomization test
    \item One study using the chi-square test
    \item One study using a Tukey test without prior ANOVA.
\end{itemize}

\subsection{Survey Files}

All papers we considered for the survey are listed in the supplementary material in the file \texttt{all\_papers.yaml} by their id in the ACL anthology bib-file. 
The 58 SDS/MDS system papers that contain new human evaluation studies and are thus included in the survey are listed in the category \texttt{with\_human\_eval}.

For the sake of completeness, we further list summarization papers we did not include in our survey. We separate them into the following categories:

\begin{description}
	\item[no\_human\_eval] 47 SDS/MDS system papers without human evaluation
	\item[sentsum] 27 Sentence summarization and headline generation papers
	\item[non\_system] 34 summarization papers that do not introduce new systems, like surveys, opinion pieces and evaluation studies
	\item[other] 10 Papers that conduct summarization with either non-textual input or non-textual output
\end{description}

We give a full list of the survey results for all papers with human evaluation studies in the file \texttt{survey\_details.csv}.
The file has the following columns:

\begin{description}
	\item[paper] Id of the paper in the ACL anthology
	\item[eval\_id] Id of the evaluation study to differentiate them in papers with multiple studies
	\item[task] Summarization task of the paper: SDS vs. MDS
	\item[genre] Genre of the summarized documents
	\item[\#docs] Number of documents in the evaluation
	\item[\#systems] Number of systems in the evaluation
	\item[includes\_reference] Whether the reference summary is included in the human evaluation
	\item[\#ann\_total] Total number of annotators in the study
	\item[\#ann\_item] Number of annotators per summary
	\item[content, fluency, repetition, coherence, referential\_clarity, other, overall] Binary columns indicating evaluation questions in the paper
	\item[measure] Annotation method used in the study
	\item[anntype] Annotator recruitment strategy
	\item[stattest] Statistical test used
	\item[design\_specified] Indicates whether it is possible to determine the full design from the information given about the study in the paper
	\item[comments] Comment column. This column describes the use of \textit{other} where present.
\end{description}

\end{document}